\documentclass[runningheads]{llncs}
\newcommand\samethanks[1][\value{footnote}]{\footnotemark[#1]}
\usepackage[T1]{fontenc}
\usepackage{graphicx}
\usepackage{marvosym}   
\usepackage{amsmath}
\usepackage{amssymb}
\usepackage{booktabs}
\usepackage{multirow}
\usepackage{xcolor}
\usepackage{footmisc}

\graphicspath{{./figure/}}

\begin{document}
\title{DINO-3DRA: Leveraging 2D Foundation Model Semantics for 3D Cerebral Aneurysm Segmentation}
\titlerunning{DINO-3DRA: 2D VFMs for 3D Aneurysm Segmentation}

\author{Jiayang Lu\inst{1} \and
Fengming Lin\inst{1,2,3}\thanks{Joint supervision.} \and
Alejandro F. Frangi\inst{1,2,3,4,5}\samethanks \and
Ali Sarrami-Foroushani\inst{1,2,3}\samethanks\thanks{Corresponding author. Email: a.sarrami@manchester.ac.uk}}
\index{Lu, Jiayang}
\index{Lin, Fengming}
\index{Frangi, Alejandro F.}
\index{Sarrami-Foroushani, Ali}
\authorrunning{J. Lu et al.}

\institute{Centre for Computational Imaging and Modelling in Medicine (CIMIM), University of Manchester, Manchester, UK \and
Christabel Pankhurst Institute, University of Manchester, Manchester, UK \and
Department of Computer Science, University of Manchester, Manchester, UK \and
Division of Informatics, Imaging \& Data Sciences, University of Manchester, Manchester, UK \and
NIHR Manchester Biomedical Research Centre, Manchester Academic Health Sciences Centre, University of Manchester, Manchester, UK \\
\email{a.sarrami@manchester.ac.uk}}
\maketitle

\begin{abstract}
Accurate aneurysm segmentation in 3D rotational angiography (3DRA) is hindered by extreme class imbalance, morphological similarity to vessels, and absent large-scale 3D pretraining. 2D vision foundation models encode dense structural priors from $\sim$1.7 billion images, yet naïve slice-wise transfer fragments anatomical continuity and destabilises optimisation. We propose DINO-3DRA, a dual-path framework achieving effective cross-dimensional semantic transfer by injecting frozen DINOv3 features into a 3D U-Net backbone via Room-Lite spatial mixing and calibrated residual fusion. On multi-centre 3DRA data, DINO-3DRA achieves state-of-the-art aneurysm segmentation (Dice: 0.758; HD95: 2.75 mm; +13\% over nnU-Net) with only 5.72M trainable parameters. Ablation studies confirm that gains arise from structured cross-dimensional transfer rather than loss design alone, with bridged foundation features improving anatomical continuity between aneurysms and parent vessels. Without fine-tuning on CADA and SHINY-ICARUS, DINO-3DRA removes all failures (Dice < 0.5) observed in baseline architectures, indicating improved stability across heterogeneous protocols. Code and weights are available at \url{https://github.com/JiayangDS/Dino3DRA}.

\end{abstract}

\section{Introduction}

Intracranial aneurysms affect 3--5\% of the population with significant
rupture mortality~\cite{vlak2011}. While 3D rotational angiography (3DRA)
enables detailed vascular imaging~\cite{vanrooij2008}, automated
segmentation remains challenging due to extreme size imbalance
($<$1\% voxels), morphological similarity to normal vessels, and
inter-scanner variability~\cite{lin2023high,lin2023adaptive}.

A fundamental asymmetry exists between 2D and 3D visual representation
learning. Self-supervised 2D vision foundation models (VFMs), particularly
DINOv3, learn dense semantic features---object boundaries, part--whole
relationships, and structural priors---through self-distillation on
$\sim$1.7 billion natural images~\cite{oquab2023,simeoni2025}. These
representations transfer effectively to medical modalities including CT
and X-ray~\cite{liu2025,xu2026}. Volumetric models, lacking comparable
large-scale 3D pretraining due to data scarcity, struggle to learn
equivalent priors~\cite{muller2025}. Rather than pursuing expensive 3D
pretraining, we leverage frozen 2D VFM features through a dual-path
architecture: the frozen branch preserves pretrained knowledge while a
trainable 3D branch learns domain-specific volumetric context, achieving
foundation-scale benefits with minimal overhead.

Several works have adapted DINO for 2D medical
imaging~\cite{xu2026,li2025,gao2025}, yet effective 2D-to-3D transfer
remains unresolved. Na\"ive slice-wise approaches fail to compete with
optimised 3D architectures~\cite{liu2025}. 

DINO-in-the-Room (DITR) projects calibrated RGB-D features onto 3D points~\cite{knaebel2025}. 3DRA has no external views, camera, or point cloud, so its 2D-to-3D problem is distinct, requiring a dense semantic volume built from slice-wise tokens. These slice-wise features, however, lack inter-slice continuity, and the natural-to-medical domain mismatch risks destabilising hybrid optimisation.

This work presents DINO-3DRA with four primary contributions:
\begin{itemize}
\begin{sloppypar}
\item \textbf{Lightweight cross-dimensional integration:} A dual-path
      architecture combining frozen DINOv3-Small with a 3D U-Net backbone~\cite{cicek2016} 
      ($\sim$120K additional parameters) achieves state-of-the-art
      aneurysm segmentation (Dice: 0.758; +13\% over nnU-Net~\cite{isensee2021}) with
      only 5.72M trainable parameters.
\end{sloppypar}      
\item \textbf{Room-Lite spatial mixer:} A parameter-efficient module
      restoring volumetric coherence from slice-wise features via
      depth-aware positional encoding and depthwise-separable convolutions.
\item \textbf{Calibrated heterogeneous fusion:} Residual injection with
      learnable scaling stabilises frozen--trainable feature alignment.
\item \textbf{Systematic validation:} Ablations confirm structured
      cross-dimensional transfer drives gains; cross-dataset evaluation
      demonstrates improved robustness, eliminating catastrophic failures
      present in all baselines.
\end{itemize}

\section{Methods}
\begin{figure}[!ht]
\centering
\includegraphics[width=\textwidth]{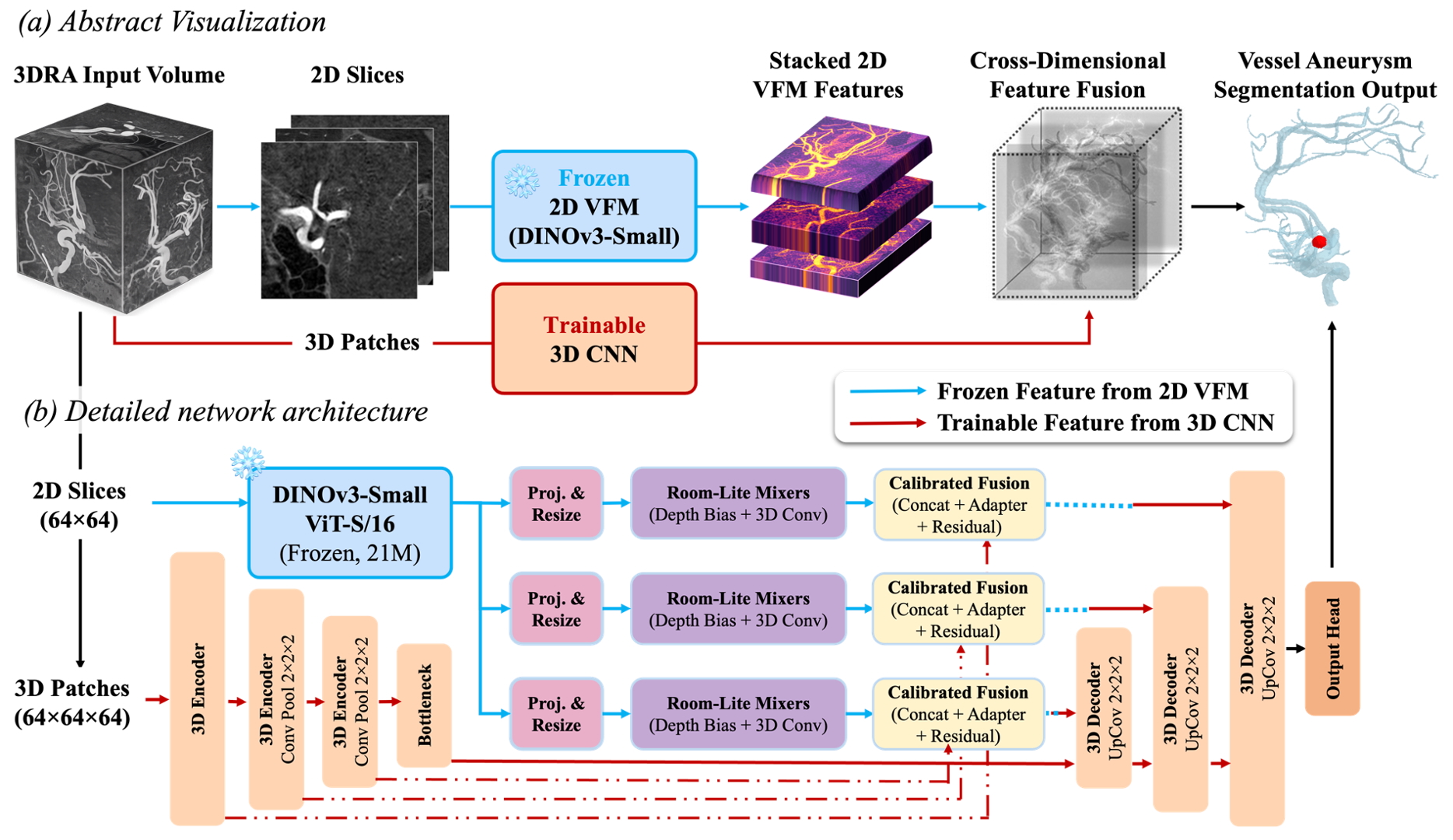}
\caption{Overview of DINO-3DRA. (a) Frozen 2D VFM (DINOv3-Small) features are fused with a trainable 3D U-Net backbone for vessel and aneurysm segmentation. (b) Multi-scale DINO features are refined by Room-Lite Mixers and fused into U-Net skip connections via Calibrated Fusion.}
\label{fig:architecture}
\end{figure}

\begin{sloppypar}
Figure~\ref{fig:architecture} illustrates DINO-3DRA's dual-path
architecture: a trainable 3D U-Net backbone~\cite{cicek2016} processing volumetric inputs and a
frozen DINOv3-Small branch extracting semantic features from 2D axial
slices, integrated via calibrated fusion modules.
\end{sloppypar}

\textbf{Dual-Path Encoder.}
The \emph{3D volumetric branch} follows the 3D U-Net 
architecture with four
encoder stages, expanding channels from 32 to 256 while downsampling
from $64^3$ to $8^3$ voxels. Each stage comprises two
$3\!\times\!3\!\times\!3$ convolutions with GroupNorm and GELU. The
symmetric decoder uses transposed convolutions and skip connections at
$64^3$, $32^3$, and $16^3$ resolutions ($\sim$5.6M parameters).

The \emph{frozen 2D semantic branch} (DINOv3-Small, ViT-S/16, 384-dim
embeddings, pretrained on LVD-1689M via self-supervised distillation)
processes all 64 axial slices independently, each replicated to RGB
and resized to $224\!\times\!224$. For each slice, the 196 patch tokens (14×14 grid) are globally averaged to a single 384-dimensional semantic embedding capturing the slice-level vascular context. These embeddings are broadcast to spatial supports at \{64², 32², 16²\},
and projected via $1\!\times\!1$ convolutions to $\{24, 48, 96\}$
channels. Stacked features are converted to isotropic 3D volumes via
trilinear interpolation.

\textbf{Room-Lite Mixers and Calibrated Fusion.}
Since DINOv3 processes slices independently, the stacked volume lacks
inter-slice continuity. Inspired by DITR~\cite{knaebel2025}, Room-Lite
mixers restore volumetric coherence via a learnable depth-positional bias
$B(z) \in \mathbb{R}^{C \times D \times 1 \times 1}$ encoding per-slice
anatomical position, and depthwise-separable 3D convolutions:
\begin{equation}
Y = X + \text{GELU}\!\left(\text{GN}_1\!\left(\text{Conv}_{1\times1\times1}
    \!\left(\text{Conv}_{3\times3\times3}\!\left(X + B(z)\right)
    \right)\right)\right).
\label{eq:roomlite}
\end{equation}
To align the statistically different DINO and U-Net feature
distributions, calibrated residual fusion is applied at each skip level:
\begin{equation}
F_{\text{fused}} = F_U + \text{GELU}\!\left(\text{GN}_1\!\left(
    W_{1\times1\times1}\left[F_U;\, \text{GN}_1(F_D)\right]\right)\right),
\label{eq:fusion}
\end{equation}
where $F_U$ and $F_D$ denote U-Net and DINOv3 features, $W$ is the
$1\!\times\!1\!\times\!1$ projection weight, and the residual connection
preserves $F_U$ when DINO features are unhelpful. 
Both equations use GN$_1$, GroupNorm with a single group, normalising over all channels and spatial positions jointly.
The decoder upsamples
via transposed convolutions, concatenates fused skip features, and
outputs three logits (background, vessel, aneurysm).

\textbf{Training Objective.}
Severe class imbalance---background ($\sim$94\%), vessels ($\sim$5\%),
aneurysms ($<$1\%)---necessitates a composite loss:
\begin{equation}
\mathcal{L} = 0.3\,\mathcal{L}_{\text{vessel}}
            + 0.5\,\mathcal{L}_{\text{aneurysm}}
            + 0.2\,\mathcal{L}_{\text{CE}}.
\label{eq:loss}
\end{equation}
Vessel Dice operates on the union of vessel and aneurysm classes to
preserve vascular continuity. Aneurysm Dice isolates lesion
segmentation~\cite{sudre2017}. Class-weighted cross-entropy
($w=\{0.1, 1.0, 2.0\}$) stabilises per-voxel classification, using
normalised inverse frequency with clamping to prevent gradient
explosion~\cite{cui2019}.

\section{Experiments}

\subsection{Dataset and Implementation}
\begin{sloppypar}
Training and evaluation used the multi-centre 3DRA @neurIST
dataset~\cite{benkner2010} (233 patients, four institutions) with expert
annotations for cerebral vessels and aneurysms, split 4:1:1
(train/val/test). Volumes were resampled to 0.35\,mm isotropic spacing,
partitioned into overlapping $64^3$ patches (stride 32), and augmented
with random rotations ($\pm15^\circ$) and flips. Case-level cross-dataset evaluation used CADA~\cite{ivantsits2020} ($n$=45, randomly sampled cases, aneurysm-only) and
SHINY-ICARUS~\cite{shinyicarus2023} ($n$=30,  randomly sampled cases, vessel+aneurysm) without
fine-tuning; these datasets differ from @neurIST in scanner uniformity,
acquisition protocols, and label completeness.
\end{sloppypar}

Training used PyTorch 2.4.0 on two NVIDIA A100(80GB) GPUs with the
Adam optimiser (lr=$1\!\times\!10^{-4}$, weight decay=$1\!\times\!10^{-5}$,
batch size 4). 





\subsection{Preliminary Experiments}

Before the main evaluation, we conducted controlled preliminary studies to select the loss function and fusion design. All runs used the same patch-level training/validation setting, with batch size 4, learning rate $1\!\times\!10^{-4}$, 60 epochs, and model selection by macro-averaged Dice. As shown in Tables~\ref{tab:loss} and~\ref{tab:arch}, the separate vessel/aneurysm Dice loss with cross-entropy achieved the best overall performance and was adopted in all subsequent experiments. For architecture design, na\"ive DINO feature concatenation degraded performance, whereas Room-Lite with calibrated fusion produced the strongest total and aneurysm Dice, supporting the need for both volumetric coherence restoration and feature-distribution alignment when transferring frozen 2D foundation features to 3D segmentation.


\begin{table}[!ht]
\centering
\caption{Loss function comparison on the baseline 3D U-Net.
         Best results in \textbf{bold}.}
\label{tab:loss}
\setlength{\tabcolsep}{4pt}
\fontsize{8}{9}\selectfont
\begin{tabular}{llccc}
\toprule
ID & Loss Type            & Dice (tot)     & Dice (ves)     & Dice (aneur) \\
\midrule
A1 & Separate Enhanced    & \textbf{0.900} & 0.927          & \textbf{0.874} \\
A2 & Simple Loss          & 0.874          & 0.919          & 0.828 \\
A3 & Asymmetric Basic     & 0.885          & 0.932          & 0.839 \\
A4 & Asymmetric Focal     & 0.857          & \textbf{0.934} & 0.780 \\
A5 & Asymmetric Adaptive  & 0.897          & 0.930          & 0.864 \\
\bottomrule
\end{tabular}
\end{table}

\begin{table}[!ht]
\centering
\caption{Architectural ablation. All DINO variants use frozen DINOv3-small.
         Best results in \textbf{bold}.}
\label{tab:arch}
\setlength{\tabcolsep}{4pt}
\fontsize{8}{9}\selectfont
\begin{tabular}{llccc}
\toprule
ID & Configuration        & Dice (tot)     & Dice (ves)     & Dice (aneur) \\
\midrule
a1 & Baseline 3D U-Net    & 0.896          & 0.922          & 0.871 \\
a2 & + DINO (naive)       & 0.885          & 0.919          & 0.850 \\
a3 & + Calib + Gating     & 0.891          & 0.924          & 0.858 \\
a4 & + Room-Lite + Calib  & \textbf{0.914} & \textbf{0.930} & \textbf{0.899} \\
b1 & Multi-Dir (learn)    & 0.859          & 0.921          & 0.798 \\
b2 & Multi-Dir (concat)   & 0.865          & 0.921          & 0.810 \\
b3 & FAPM Projection      & 0.866          & 0.923          & 0.809 \\
\bottomrule
\end{tabular}
\end{table}

\textbf{Post-Processing.} Overlapping patch predictions are aggregated 
via majority voting. The aneurysm threshold (0.005) was selected by grid 
search over $[0, 0.1]$ on the baseline and applied uniformly, maximising 
aneurysm Dice with ${>}90\%$ voxel-level recall; vessel threshold follows 
the standard 0.5. Candidates are scored by $S = P_{\max} + \log_{10}(V)$ 
and filtered by minimum size ($\geq$40 voxels) and vessel proximity.

\subsection{Comparison with State-of-the-Art}

\begin{figure}[!ht]
\centering
\includegraphics[width=\textwidth]{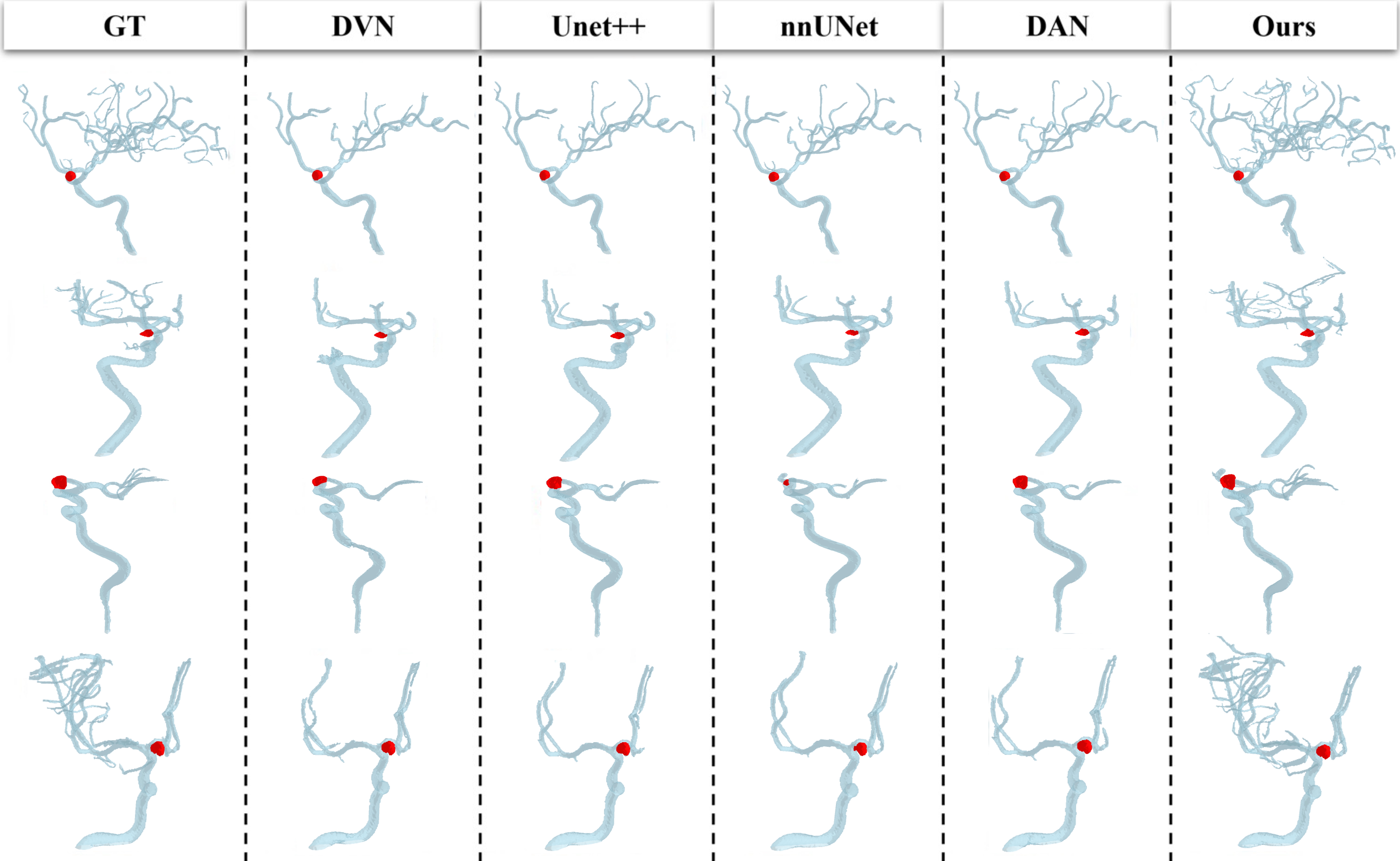}

\caption{Qualitative comparison (Blue: Vessel / Red: Aneurysm).
         Left to right: Ground truth (GT), DeepVesselNet (DVN),
         U-Net++, nnU-Net, Dual Attention (DAN), DINO-3DRA (Ours).}
\label{fig:qualitative}
\end{figure}

Table~\ref{tab:sota} compares DINO-3DRA with established baselines
(paired Wilcoxon tests, $p<0.05$). DINO-3DRA achieves the strongest
aneurysm performance (Dice: 0.758, +13\% over nnU-Net) with notably
lower variance ($\sigma=0.033$ vs.\ 0.097--0.138 for baselines).
Qualitative results (Fig.~\ref{fig:qualitative}) show clearer aneurysm
boundaries and improved vascular continuity frequently missed by
competing methods.

\begin{table}[!ht]
\centering
\caption{Quantitative comparison on @neurIST dataset. $p$-values
         vs.\ Ours (paired Wilcoxon signed-rank test). Best results in \textbf{bold}.}
\label{tab:sota}
\setlength{\tabcolsep}{3pt}
\fontsize{8}{9}\selectfont
\begin{tabular}{llcccc}
\toprule
Method & Structure & Dice & Jaccard & Vol.\ Sim. & $p$ \\
\midrule
\multirow{2}{*}{U-Net++ ($\sim$9M)~\cite{zhou2019}}
  & Vessel   & 0.820$\pm$0.109 & 0.827$\pm$0.123 & 0.946$\pm$0.100 & $<$0.01 \\
  & Aneurysm & 0.598$\pm$0.340 & 0.504$\pm$0.322 & 0.658$\pm$0.337 & $<$0.01 \\
\midrule
\multirow{2}{*}{Dual Attention ($\sim$18M)~\cite{fu2019}}
  & Vessel   & 0.889$\pm$0.138 & 0.818$\pm$0.145 & 0.945$\pm$0.095 & $<$0.01 \\
  & Aneurysm & 0.595$\pm$0.337 & 0.499$\pm$0.321 & 0.663$\pm$0.334 & $<$0.01 \\
\midrule
\multirow{2}{*}{DeepVesselNet ($\sim$2M)~\cite{tetteh2020}}
  & Vessel   & 0.892$\pm$0.097 & 0.815$\pm$0.117 & 0.946$\pm$0.074 & $<$0.01 \\
  & Aneurysm & 0.156$\pm$0.235 & 0.108$\pm$0.190 & 0.262$\pm$0.277 & $<$0.01 \\
\midrule
\multirow{2}{*}{nnU-Net ($\sim$31M)~\cite{isensee2021}}
  & Vessel   & 0.869$\pm$0.120 & 0.782$\pm$0.137 & 0.913$\pm$0.107 & $<$0.01 \\
  & Aneurysm & 0.669$\pm$0.319 & 0.575$\pm$0.306 & 0.732$\pm$0.307 & $<$0.01 \\
\midrule
\multirow{2}{*}{\textbf{Ours} ($\sim$5.72M)}
  & Vessel   & \textbf{0.897}$\pm$0.033 & \textbf{0.815}$\pm$0.053 & \textbf{0.962}$\pm$0.036 & --- \\
  & Aneurysm & \textbf{0.758}$\pm$0.234 & \textbf{0.661}$\pm$0.253 & \textbf{0.838}$\pm$0.239 & --- \\
\bottomrule
\end{tabular}
\end{table}

\subsection{Ablation Study}

\begin{figure}[!ht]
\centering
\includegraphics[width=\textwidth]{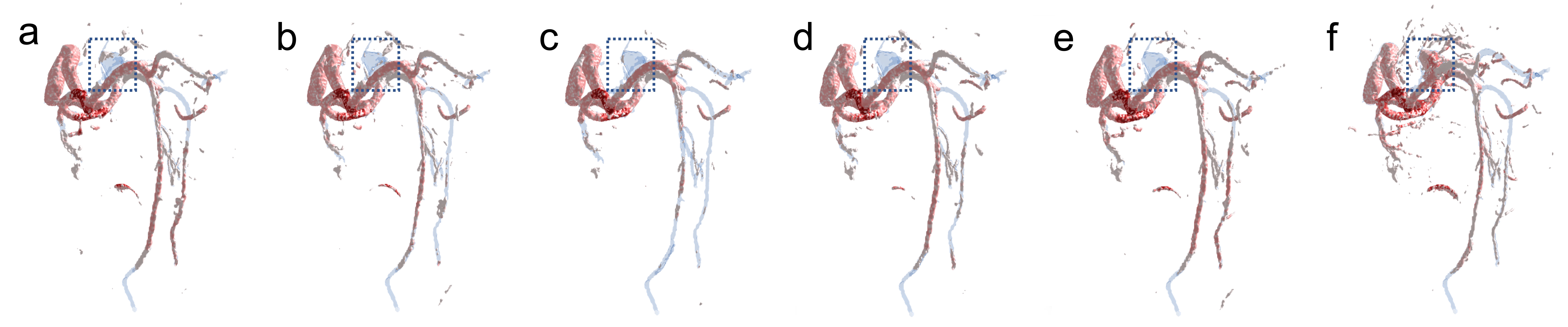}
\caption{Vessel confidence maps for ablation variants. Ground truth
(semi-transparent blue), predictions (red; darker = higher confidence).
(a)~Baseline U-Net, (b)~Random Features, (c)~w/o Room-Lite,
(d)~w/o Calibration, (e)~DINOv2-small, (f)~DINO-3DRA (Ours).}
\label{fig:ablation}
\end{figure}

We quantified the individual contributions of semantic transfer,
volumetric coherence restoration, and calibrated fusion under identical
training settings (Table~\ref{tab:ablation}). Compared configurations:
(i)~Baseline 3D U-Net (5.60M); (ii)~Full DINO-3DRA (5.72M);
(iii)~Random-feature control, replacing pretrained DINOv3 with a
ViT-S/16 using standard initialisation (truncated normal, std=0.02);
(iv)~w/o Room-Lite, directly stacking 2D features without spatial mixing;
(v)~w/o Calibration, using na\"ive concatenation in place of calibrated
residual fusion; and (vi)~DINOv2-small substitution.
\begin{table}[!ht]
\centering
\caption{Ablation results (paired Wilcoxon signed-rank test; HD95 in mm).}
\label{tab:ablation}
\setlength{\tabcolsep}{4pt}
\fontsize{8}{9}\selectfont
\begin{tabular}{lcccccc}
\toprule
\multirow{2}{*}{Configuration}
  & \multicolumn{3}{c}{Aneurysm}
  & \multicolumn{3}{c}{Vessel} \\
\cmidrule(lr){2-4}\cmidrule(lr){5-7}
  & Dice & $p$ & HD95 & Dice & $p$ & HD95 \\
\midrule
Ours (Full)     & \textbf{0.758}$\pm$0.234 & ---     & \textbf{2.75} & \textbf{0.897}$\pm$0.033 & ---     & \textbf{3.74} \\
w/o Room-Lite   & 0.385$\pm$0.310          & $<$.001 & 22.08         & 0.857$\pm$0.043          & $<$.001 & 5.64 \\
w/o Calibration & 0.492$\pm$0.301          & $<$.001 & 10.12         & 0.857$\pm$0.045          & $<$.001 & 6.71 \\
Random Features & 0.628$\pm$0.279          & $<$.001 &  6.15         & 0.885$\pm$0.039          & .042    & 4.41 \\
DINOv2-small    & 0.590$\pm$0.252          & $<$.001 &  6.50         & 0.868$\pm$0.050          & $<$.001 & 4.30 \\
Baseline U-Net  & 0.711$\pm$0.240          & .030    &  4.20         & 0.873$\pm$0.046          & .002    & 5.41 \\
\bottomrule
\end{tabular}
\end{table}
Figure~\ref{fig:ablation} visualises vessel confidence maps for a
challenging case. Variants w/o Room-Lite, w/o Calibration, and Random
Features all exhibit low confidence across the aneurysm dome
(Dice: 0.385--0.628), failing to recognise vascular tissue. Baseline
U-Net and DINOv2 achieve reasonable detection but display fragmented
confidence, treating aneurysms as isolated entities. Full DINO-3DRA
produces homogeneous confidence across the entire aneurysm
(Dice: 0.758), recognising aneurysms as pathological vessel dilations
rather than separate anomalies---anatomical consistency that provides a better geometric starting point for downstream surface meshing and hemodynamic modeling~\cite{lin2024transwarp,lin2024gsema}.

\subsection{Cross-Dataset Generalization}

\begin{figure}[!ht]
\centering
\includegraphics[width=\textwidth,height=0.40\textheight,keepaspectratio]{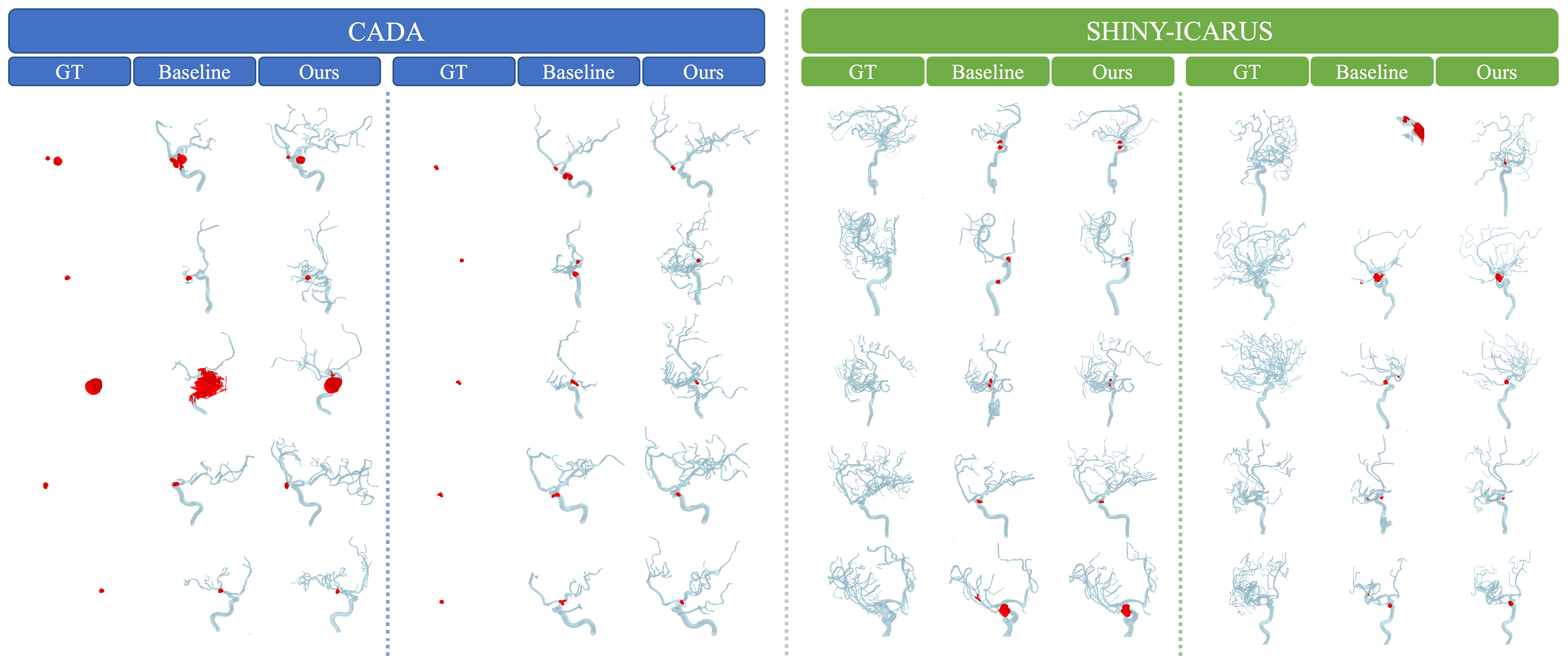}
\caption{Cross-dataset generalization validation on CADA (left) and SHINY-ICARUS
(right). Blue: vessel; Red: aneurysm. Columns show ground truth,
baseline, and DINO-3DRA predictions.}
\label{fig:crossdataset}
\end{figure}

Models trained on @neurIST were evaluated on external datasets without
fine-tuning: CADA ($n$=45) and SHINY-ICARUS ($n$=30). Results are
summarised in Table~\ref{tab:crossdataset} and Fig.~\ref{fig:crossdataset}.
On SHINY-ICARUS, DINO-3DRA achieved Dice +10.4\% ($p<$0.001) with HD95
reduced by 75.8\% (15.52$\to$3.76\,mm). On CADA, mean Dice improved
modestly ($p$=0.607) but variance halved ($\sigma$: 0.186$\to$0.093).
Critically, DINO-3DRA eliminated all failures (Dice < 0.5) on both
datasets (CADA: 11.1\%$\to$0\%; SI: 3.3\%$\to$0\%), reducing the
coefficient of variation by 52--70\%.
\begin{table}[!ht]
\centering
\caption{Cross-dataset generalization results. Failures: Dice$<$0.5;
         SI denotes SHINY-ICARUS. Best results in \textbf{bold}.}
\label{tab:crossdataset}
\setlength{\tabcolsep}{3pt}
\fontsize{8}{9}\selectfont
\begin{tabular}{llccccc}
\toprule
Dataset & Model & Dice & Jaccard & Vol.\ Sim. & HD95 & Failures \\
\midrule
\multirow{2}{*}{CADA}
  & Baseline  & 0.711$\pm$0.186          & 0.577$\pm$0.186          & 0.813$\pm$0.128 &  5.16 & 11.1\% \\
  & DINO-3DRA & \textbf{0.739}$\pm$0.093 & \textbf{0.594}$\pm$0.118 & 0.807$\pm$0.109 & \textbf{3.58} & \textbf{0\%} \\
\midrule
\multirow{2}{*}{SI}
  & Baseline  & 0.718$\pm$0.147          & 0.574$\pm$0.129          & 0.808$\pm$0.089 & 15.52 &  3.3\% \\
  & DINO-3DRA & \textbf{0.793}$\pm$0.049 & \textbf{0.660}$\pm$0.065 & 0.812$\pm$0.056 & \textbf{3.76} & \textbf{0\%} \\
\bottomrule
\end{tabular}
\end{table}
Case analysis reveals that DINO-3DRA tends to under-segment large or
morphologically atypical aneurysms, reflecting the model's strong vessel
prior: ambiguous boundary voxels at the dome periphery (where intensity
gradients are subtle) are conservatively assigned to the vessel class.
Two factors amplify this effect: (1)~training data consisting
predominantly of compact saccular aneurysms, while CADA includes fusiform
morphologies with inherently ambiguous boundaries; and (2)~large aneurysms
spanning multiple patches, where boundary consistency degrades. This trade-off favours clinical utility: reliable localisation and reduced failure rates over volumetric completeness, where consistent detection matters more than precise boundary delineation.

\section{Conclusion}

This paper presents DINO-3DRA, a framework that transfers frozen 2D
foundation model features to 3D medical segmentation via Room-Lite
spatial mixers and calibrated residual fusion. Ablation studies confirm
that proper bridging mechanisms are essential for cross-dimensional
semantic transfer, yielding anatomically coherent predictions in which
aneurysms are recognised as pathological vessel dilations. The method
achieves state-of-the-art aneurysm segmentation (Dice: 0.758,
+13\% over nnU-Net) and vessel segmentation (Dice: 0.897) with
substantially reduced prediction variance. Cross-dataset evaluation
eliminates all failures cases (Dice < 0.5) while preserving fine vascular
structures and vessel-aneurysm continuity, suggesting improved robustness across heterogeneous protocols, to be confirmed on larger external cohorts.


\begingroup
\emergencystretch=2em
\subsubsection{Acknowledgements.}
AFF acknowledges support from the Royal Academy of Engineering under the RAEng Chair in Emerging Technologies
(INSILEX CiET191919), ERC Advanced Grant--UKRI Frontier Research Guarantee
(INSILICO EPY0304941), the UK Centre of Excellence on in-silico Regulatory Science and Innovation
(UK CEiRSI) (10139527), the National Institute for Health and Care Research (NIHR) Manchester Biomedical Research Centre
(BRC) (NIHR203308), the BHF Manchester Centre of Research Excellence
(RE2430017), and the CRUK RadNet Manchester
(C1994A28701).
\par
\endgroup

\subsubsection{Disclosure of Interests.}
AF is Co-founder and shareholder of OculomeX Health Ltd, adsilico Ltd, and Synaptive Consulting Ltd.

\bibliographystyle{splncs04}
\bibliography{Paper-2993}

\end{document}